\documentclass[conference]{IEEEtran}
\usepackage{graphicx}
\usepackage{amsmath}
\usepackage{booktabs}
\usepackage{subfig}
\usepackage{multirow}
\usepackage{algorithm}
\usepackage[noend]{algpseudocode}
\IEEEoverridecommandlockouts

\begin{document}
\title{Measuring Agreement on Linguistic Expressions in Medical Treatment Scenarios}
\author{\IEEEauthorblockN{J. Navarro$ ^{a} $, C. Wagner$ ^{a} $, U. Aickelin$ ^{b} $, L. Green$ ^{c} $, R. Ashford$ ^{c} $}
\IEEEauthorblockA{$ ^{a} $Lab for Uncertainty in Data and Decision Making (LUCID) and Horizon,\\School of Computer Science, University of Nottingham, Nottingham, UK\\$ ^{b} $University of Nottingham, Ningbo, China\\
Nottingham University Hospitals, UK\\
Email: \{psxfjn,christian.wagner,uwe.aickelin\}@nottingham.ac.uk\\Email: lynsey.green@nuh.nhs.uk, rashford@nhs.net}\thanks{This work was partially funded by the RCUK grant EP/M02315X/1 From Human Data to Personal Experience.}}
\maketitle

\begin{abstract} 
Quality of life assessment represents a key process of deciding treatment success and viability. As such, patients' perceptions of their functional status and well being are important inputs for impairment assessment. Given that patient completed questionnaires are often used to assess patient status and determine future treatment options, it is important to know the level of agreement of the words used by patients and different groups of medical professionals. In this paper, we propose a measure called the Agreement Ratio which provides a ratio of overall agreement when modelling words through Fuzzy Sets (FSs). The measure has been specifically designed for assessing this agreement in fuzzy sets which are generated from data such as patient responses. The measure relies on using the Jaccard Similarity Measure for comparing the different levels of agreement in the FSs generated. Synthetic examples are provided in order to show how to calculate the measure for given Fuzzy Sets. An application to real-world data is provided as well as a discussion about the results and the potential of the proposed measure.
\end{abstract}

\IEEEpeerreviewmaketitle

\section{Introduction}
In the context of medical treatment, capturing patients' and medical professionals' perceptions of functional status is an important instrument to consider when evaluating possible outcomes after treatment intervention (e.g., job modifications, use of assistive devices, etc.) \cite{Davis1996}. In this context, it is important to be aware of the uncertainty associated to the words (linguistic descriptors) used by the stakeholders (e.g., physiotherapists, surgeons, patients, etc.): this includes variability in people's perceptions throughout the day, experience, professional background, etc.

In \cite{Zadeh1996}, Zadeh introduced the Computing With Words (CWW) paradigm in which, according to him, words and propositions from natural language are used as objects of computation. As such, this paradigm focuses on narrowing the differences between human reasoning and computing by allowing the manipulation of different and usually imprecise perceptions \cite{Herrera2009}.\\
Surveying groups of people enables the capture of uncertainty through intervals on areas of interest, which is an important resource for capturing the variations of perceptions among those surveyed. A number of methods have been developed in order to capture stakeholder perceptions of words/concepts expressed through interval-based surveys, including \cite{Wagner2014,Coupland2010,Feilong2008a}. Basically, they rely on allowing participants to express their uncertainty about a given response by providing an interval. Such participants' intervals are subsequently used to generate a Fuzzy Set model (which depends of the method employed and the types of uncertainty being modelled) representing the overall perception (or a subset of the participants) of the initial word surveyed.

There are a number of measures for Type-1 Fuzzy Set (T1 FS) agreement models which have shown to be useful for several purposes. Similarity measures \cite{Setnes1998} for example, are functions which indicates the degree to which two FSs are similar. The Jaccard similarity measure \cite{Jaccard1908}, has been applied to both relate and compare word models to concept models in different contexts \cite{Wagner2013b,Navarro2016a}. In \cite{Miller2014}, an exploration of attributes (e.g., \textit{Support Size}, \textit{Height}, \textit{Spread}, \textit{Core Size} and \textit{Fuzziness}) obtained from T1 FSs agreement models was performed. Such exploration found that additional information related to the consensus can be extracted with regard to traditional statistical measures. However, a direct measure of agreement among participants expressing how well conceived a given word is in an specific context where imprecise descriptors are being used as a basis to assess patients' health conditions has not been reported yet.

In this paper, we focus on presenting an Agreement Ratio measure which aims to provide a measure for the inter-participant agreement on perception of words (linguistic descriptors) using T1 FSs derived from interval-based surveys. This paper is structured as follows. Section \ref{sec:back} provides background on a questionnaire called Toronto Extremity Salvage Score (TESS) for the assessment of impairment in which linguistic descriptors are a key element for expressing patients' perceptions, T1 FSs and modelling of inter-participant uncertainty using the Interval Agreement Approach. Section \ref{sec:agr} introduces the motivation behind the proposal of this measure and details about its practical implementation. Section \ref{sec:res} presents a series of numeric examples for different data followed by the application to real world data obtained from patients and medical professionals using an interval-valued questionnaire. Finally, Section \ref{sec:disc} provides a discussion of the results presented and the application of the proposed measure on different types of T1 FSs, while Section \ref{sec:conc} presents the conclusions and future work derived from this proposed measure.

\section{Background}\label{sec:back}
The following introduces the TESS questionnaire used to assess patient' functional status, which will serve as the context for the experiments presented later in the paper. It is followed by a brief overview of T1 and Interval Type-2 FSs and two methods employed to model the agreement among a group of stakeholders using FSs. Both approaches will be used in the paper to generate FSs from data which in turn will be evaluated using the proposed measure.
\subsection{Toronto Extremity Salvage Score (TESS)}
\begin{figure}[h]
\centering
\includegraphics[scale=0.4]{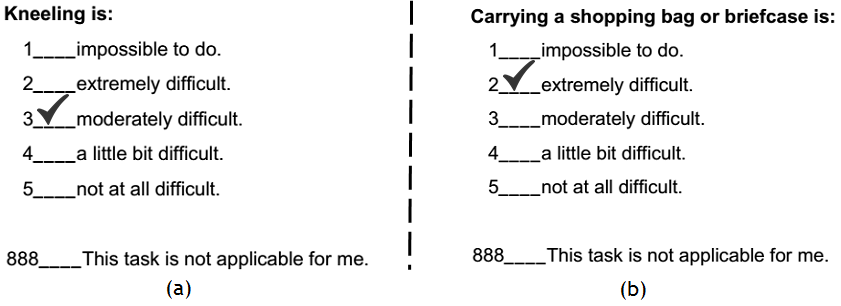}
\caption{Two sample TESS items: (a) item taken from the lower extremity questionnaire, (b) item taken from the upper extremity questionnaire.}
\label{fig:tess}
\end{figure}

The Toronto Extremity Salvage Score (TESS) is a disease-specific measure developed for patients undergoing limb preservation surgery for tumours of the extremities \cite{Davis1996}. It is a patient-completed questionnaire with questions framed to ask about the difficulty experienced performing daily activities over the last week aimed to monitor the effects of therapeutic interventions. TESS is commonly administered at four time points: the first session (which is commonly before surgery) and 12, 18 and 24 months from then on. The TESS consists of 30 and 29 items for lower limb and upper limb cases respectively with items such as the the one shown in Fig. \ref{fig:tess}.
As can be seen, difficulty is rated on a 5-point Likert-type scale ranging from ``\textit{not at all difficult}'' to ``\textit{impossible to do}''. Commonly, after having been completed by the patient, the whole set of answers is used to generate a standardized score ranging from 0 to 100. This evaluated TESS is finally analysed by surgeons/physiotherapists in order to measure changes in physical functions over time.

\subsection{Type-1 Fuzzy Sets}%
Fuzzy Sets are sets in which, unlike in traditional set theory, the membership of each element is a number in the interval $[0,1]$. Given a universe of discourse $X$, a FS $A$  is represented as a set of ordered pairs of an element $x$ and its membership value within $A$, denoted by $ {\mu}_{A}(x) $, i.e.
\begin{equation}
A={(x,{\mu}_{A}(x))| x\in X}
\end{equation}
Alpha-cuts (or $\alpha$-cuts) are an important concept in FSs, given that a FS $A$ can also be represented as a collection of its $\alpha$-cuts \cite{zadeh1975concept}. An $\alpha$-cut of a FS $A$ is a crisp set defined as
\begin{equation}
{A}_{\alpha}=\{{x}|{\mu}_{A}(x)\ge \alpha, \alpha\in [0,1]\}
\end{equation}
\subsection{Interval Agreement Approach}
The Interval Agreement Approach (IAA) was introduced in \cite{Wagner2014} as a method for generating FSs from surveys in which answers are given as interval-valued data representing uncertainty in people's opinions/perceptions. It is built on top of the work presented in \cite{Miller2012}, where an agreement-based method \cite{Wagner2011} of capturing interval-valued survey data is demonstrated.

The IAA considers two types of intervals in the process of capturing responses: crisp (no uncertainty about the interval endpoints) and uncertain (each endpoint modelled itself as a crisp interval). It considers two types of uncertainty to be modelled through different dimensions of the resultant FSs, namely inter-source (variation among a group of participants) and intra-source (variation in the opinion of a particular participant). Depending on the data, the IAA can generate:
\begin{itemize}
	\item Type-1 FSs. When crisp intervals and either inter- or intra-source uncertainty are modelled in the primary degree of membership by combining multiple intervals,
	\item Interval Type-2. When uncertain intervals and also, either inter- or intra-source uncertainty is modelled in the primary degree of membership by combining multiple intervals,
	\item General Type-2 FS based on zSlices \cite{Wagner2010a}. In this case, both inter- and intra-source uncertainty are being modelled through the primary and secondary degrees of membership.
\end{itemize}
 In this paper, we are focusing on the agreement ratio among stakeholders (inter-participant uncertainty). Such uncertainty is captured through crisp intervals and consequently, it is modelled by employing the IAA to generate T1 FSs. We provide a brief review of generating T1 FSs using the IAA below:

Consider $ N $ (closed) intervals $ { \bar { A }  }_{ i } =[{ l }_{ \bar { { A }_{ i } }  },{ r }_{ \bar { { A }_{ i } }  }]$, $ i\in \left\{ 1,...,N \right\} $ to be modelled as a T1 FS $ A $ where the intervals are delimited by ${ l }_{ \bar { { A }_{ i } }  }$ and ${ r }_{ \bar { { A }_{ i } }  }$. The membership function of $ A $ (denoted by ${\mu}_{A}$) given in (\ref{eq:iaaSets}).

\begin{equation}\label{eq:iaaSets}
\begin{aligned}
\mu (A)={} & { { y }_{ 1 } }/{ \bigcup _{ { i }_{ 1 }=1 }^{ N }{ { \bar { A }  }_{ { i }_{ 1 } } }  }\\
& +{ { y }_{ 2 } }/\left( { \bigcup _{ { i }_{ 1 }=1 }^{ N-1 }{ \bigcup _{ { i }_{ 2 }={ i }_{ 1 }+1 }^{ N }{ \left( { \bar { A }  }_{ { i }_{ 1 } }\cap { \bar { A }  }_{ { i }_{ 2 } } \right)  }  }  } \right) \\
&+{ { y }_{ 3 } }/\left( { \bigcup _{ { i }_{ 1 }=1 }^{ N-2 }{ \bigcup _{ { i }_{ 2 }={ i }_{ 1 }+1 }^{ N-1 }{ \bigcup _{ { i }_{ 3 }={ i }_{ 2 }+1 }^{ N }{ \left( { \bar { A }  }_{ { i }_{ 1 } }\cap { \bar { A }  }_{ { i }_{ 2 } }\cap { \bar { A }  }_{ { i }_{ 3 } } \right)  }  }  }  } \right) \\
&+\cdots \\
& +{ { y }_{ N } }/\left( { \bigcup _{ { i }_{ 1 }=1 }^{ 1 }{ \cdots \bigcup _{ { i }_{ N }=N }^{ N }{ \left( { \bar { A }  }_{ { i }_{ 1 } }\cap \dots \cap { \bar { A }  }_{ { i }_{ N } } \right)  }  }  } \right) ,
\end{aligned}
\end{equation}
where ${y}_{i}=\frac{i}{N}$and $/$ refers to the common notation of membership, not division. For practical applications, (\ref{eq:iaaSets}) can be calculated in a recursive and discrete manner by formulating the function as
\begin{equation}\label{eq:IAA}
\begin{aligned}
	{ \mu  }_{ A }\left( x' \right) =\frac { \left( \sum _{ i=1 }^{ N }{ { \mu  }_{ \bar { { A }_{ i } }  }\left( x' \right)  }  \right)  }{ N } ,
	\end{aligned}
\end{equation}

\begin{center}
	where: $ { \mu  }_{ \bar { { A }_{ i } }  }\left( x'  \right) =\begin{cases} 1\qquad { l }_{ \bar { { A }_{ i } }  }\le x' \le { r }_{ \bar { { A }_{ i } }  } \\ 0\qquad else \end{cases}. $
\end{center}

\section{Agreement Ratio}\label{sec:agr}
This section proposes a method of generating a useful value for the analysis of linguistic information represented as FSs, called the Agreement Ratio. Section \ref{sec:mot} reviews the aims and motivation behind the proposed agreement ratio, followed in Section \ref{sec:met} by an in-depth description of the measure.

\subsection{Motivation}\label{sec:mot}
The use of words as a means of communication between patient-medical staff is a natural way of expressing perceptions. However, the challenge of dealing with different interpretations of words (``words mean different things to different people'') leads to looking for a standardised vocabulary, as has been suggested by the European Union \cite{Berry2003}. Therefore, there is a need for a means to analysing how similar the understanding of key words is across stakeholders in patient treatment.

\begin{figure}[!ht]
\centering
	\subfloat[\label{subfig:joint}]{
		\includegraphics[width=0.26\textwidth]{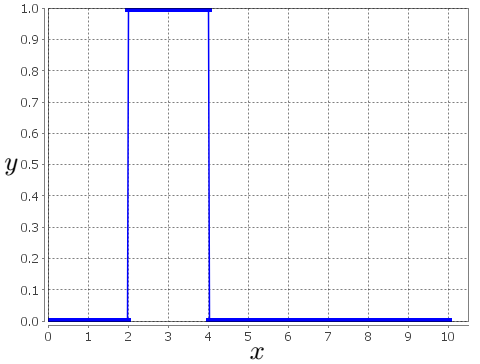}
	}\hfill
	\subfloat[\label{subfig:disjoint}]{
		\includegraphics[width=0.26\textwidth]{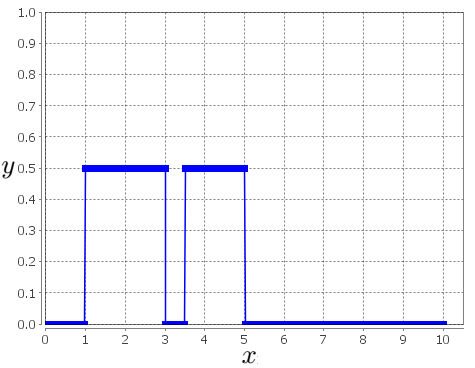}
	}\hfill
	\caption{FSs generated from the intervals \({ \overline { A }  }_{ 1 }=\left[ 2,4 \right] \) and \({ \overline { A }  }_{ 2 }=\left[ 2,4 \right] \) for Fig. (\ref{subfig:joint}), and the intervals \({ \overline { A }  }_{ 1 }=\left[ 1,3 \right] \) and \({ \overline { A }  }_{ 2 }=\left[ 3.5,5 \right] \) for Fig. (\ref{subfig:disjoint}).}
\label{fig:jointDisjoint}
\end{figure}

The aim of the agreement ratio is to provide a number contained within the interval $[0,1]$ representing the extent of agreement among a group of surveyed stakeholders (without discarding particular responses) whose responses are modelled in the given FS. Therefore, a method for generating data-driven FS models, considering the stakeholders opinions is key. As such, we have considered FSs generated with the IAA and EIA methods to analyse the agreement obtained.\\
Two key assumptions for the measure are:
\begin{enumerate}
\item \textit{Agreement} is when 2 or more sources coincide in a given point/value/opinion.
\item The more opinions overlap at a particular region, the stronger the agreement is conceived.
\end{enumerate}
Initially, we began developing the measure considering that in the IAA given the inter-participant variation is reflected directly in the primary degree of membership $ y $ (or $ \mu $). To illustrate, consider two simple contrasting cases in which two intervals/participants ($N=2$) are used to create a FS:
\begin{itemize}
\item Two identical intervals. An agreement ratio must be equal to 1 given that all stakeholders (intervals) totally agree (overlap).
\item Two disjoint intervals. An agreement ratio must be equal to 0 given that there are no regions in which the stakeholders agree (Fig. (\ref{subfig:disjoint})).
\end{itemize}

From these initial cases, it can be seen that at the highest level of membership (we will refer to it as $y_N$), there might be one or more intervals representing regions where the $N$ intervals agree, at the ${y}_{N-1}$ level, there might be one or more overlapping regions where at least $N-1$ intervals agree and so on. Thus, if a proportion of each of the $y$ levels contained with the next lower level is calculated, then a ratio representing their quantitative relation.
For example, in the FS depicted in Fig. \ref{subfig:joint}, the length of the agreement interval at the $y_2$ level is 2, which is equal to the one at the $y_1$ level and thus, the relation can be represented as $\frac{2}{2}=1$. For the FS of figure \ref{subfig:disjoint}, such relation between the length at level $y_2$ is 0 since there is not any region where both intervals overlapped and the length at level $y_1$ is $2+1.5=3.5$ can be represented as $\frac{0}{3.5}=0$.
Moreover, considering cases with more intervals to analyse, regions with higher agreement over others with less must contribute to the ratio with ``higher relevance''.

Using these cases as basis we can proceed to generalise and propose a method for calculating an agreement ratio for a FS in Section \ref{sec:met}.

\subsection{Method}\label{sec:met}
Let ${ \bar { A }  }_{ n }$, $n\in\left\{ 1,...,N \right\} $ be a set of intervals ${ \bar { A }  }_{ i }=\left\{ { l }_{ { \bar { A }  }_{ i } },{ r }_{ { \bar { A }  }_{ i } } \right\} $. The IAA uses the set of intervals to model the overall agreement through a FS based model where the membership value of each $x\in X$ accounts for the ratio of a given x contained in the set of intervals.\\

An agreement ratio $\gamma$ is obtained from a T1 FS with the following equation:

\begin{equation}\label{eq:agrRat}
\gamma \left( A \right) ={ \left( { y }_{ N }\left( \frac { { \left| \bar { A }  \right|  }_{ N } }{ { \left| \bar { A }  \right|  }_{ N-1 } }  \right) +\cdots +{ y }_{ 2 }\left( \frac { { \left| \bar { A }  \right|  }_{ 2 } }{ { \left| \bar { A }  \right|  }_{ 1 } }  \right)  \right)  }/{ \sum _{ i=2 }^{ N }{ { y }_{ i } }  }
\end{equation}
where $ 0\le \gamma\le 1$, $/$ refers to division and ${y}_{i}=\frac { i }{ N }$ weights the relation between immediate agreement $y$ levels in question. It can be noticed that the lowest level $y_1$ is not being used because the agreement is conceived when 2 or more intervals overlap. Also, we use ${ \left| \bar { A }  \right|  }_{ i }$ to represent the total length(s) of the set(s) of intervals with all possible \textit{i}-tuple intersection of intervals associated to the $y_i$ agreement level. For example, the length ${ \left| \bar { A }  \right|  }_{ 1 }$ is equal to the length of the support of $A$ since it is the union of all intervals whereas ${ \left| \bar { A }  \right|  }_{ N }$ is equal to the length of the intersection of all intervals. Finally, the overall summation is divided by the sum of ``weights'' so the final ratio is normalised to a number in the range [0,1].

The term ${ \left| \bar { A }  \right|  }_{ N }$ can be represented as described in (\ref{eq:length}).
\begin{equation}\label{eq:length}
{ \left| \bar { A }  \right|  }_{ N }=\sum _{ { i }_{ 1 }=1 }^{ 1 }{ \cdots \sum _{ { i }_{ N }=N }^{ N }{ \left| { \bar { A }  }_{ { i }_{ 1 } }\cap \dots \cap { \bar { A }  }_{ { i }_{ N } } \right|  }  } 
\end{equation}
Considering that such calculations can involve handling a considerable number of combinations to compute as the number of participants/intervals increases, (\ref{eq:length}) can be estimated through "discretisations" in practical applications by using alpha cuts (instead of the so-called $y$ agreement levels) such as described in Algorithm\ref{alg:alpha}. A consideration for the calculation of the $\gamma$ measure using alpha-cuts is: if IAA generated FS models are used, then the number of alpha cuts can be chosen to be equal to the number of intervals/participants; if any other method is used, then it depends of the number of desired "discretisations".
\begin{algorithm}
\caption{Estimation of lengths based on $\alpha$-cuts}\label{alg:alpha}
\begin{algorithmic}[1]
\Procedure{AlphaLength}{$\alpha$, $A$}\Comment{The sum of lengths}
\State $l\gets 0$, $r\gets 0$
\State $ N\gets $ \# of discretisations
\State discretise($x$)\Comment{discretise domain $x_1,...,x_i,...x_N$}
\State $b\gets $false\Comment{Boolean for detection of intervals}
\For{$i=1$ to $N$}
\State $y_i\gets {\mu}_{A}(x_i)$
\If{$y_i<\alpha$}
\State $y_i\gets 0$
\If{b = true}
\State $r\gets {x}_{i-1}$
\State addCut$(l,r)$\Comment{Add the detected interval}
\EndIf
\State $b\gets$ false
\Else
\If {$b=$false}
\State $l\gets{x}_{i}$
\EndIf
\State $b\gets true$
\EndIf
\EndFor
\If {$b= true$}
\State $r\gets x_{N}$
\State addCut$(l,r)$
\EndIf
\For{each $\alpha$Cut $j$}
\State $length\gets length + (r_{j}-l_{j})$\Comment{Interval Size}
\EndFor
\State \textbf{return} $length$
\EndProcedure
\end{algorithmic}
\end{algorithm}
It should be noticed that, the use of alpha cuts can allow the Agreement Ratio to be applied in any T1 FS regardless the method employed to generate it from intervals. However, as stated in the motivation, the degree of membership of the FS in question is assumed to express the extent of agreement among the surveyed stakeholders. This assumption, allows the proposed measure to take advantage of different FSs (normal or non-normal, convex or non-convex ) from an interpretative point of view, in order to show so, we will analyse different FSs shapes and provide the results in next section.
\section{Results}\label{sec:res}
In this section, we present synthetic examples of application of the proposed agreement ratio considering diverse types of T1 MFs and finally, its application to real-world data obtained from three groups of people involved in a pilot study. For the real-world application, we present the results of using models obtained from the IAA.

\subsection{Synthetic Examples 1 using convex FSs}
\begin{figure}[h]
\includegraphics[scale=0.4]{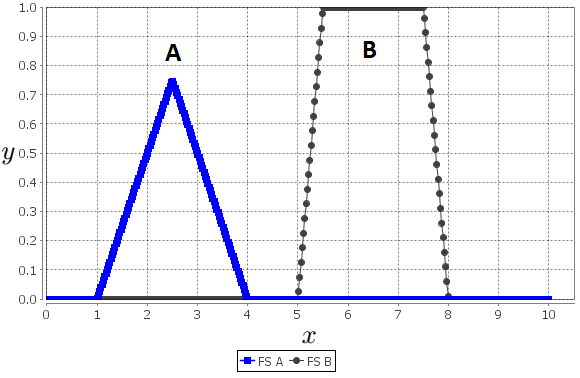}
\centering
	\caption{{\fontsize{8pt}{1em} \selectfont Two convex Fuzzy sets \textit{A} (non-normal) and \textit{B} (normal).}}
	\label{fig:convex}
\end{figure}

Consider the FSs depicted in Fig. \ref{fig:convex} where no assumptions about the method used to generate them has been made other than, that the membership axis represents the level of agreement among the participants. Thus, one of the FSs depicted has been chosen to be non-normal deliberately so values from the calculated Agreement ratio can be contrasted. For comparison purposes, we also consider two common shapes for FS membership functions: triangular and trapezoidal.
For the FS \textit{A}, the calculated agreement ratio using 10 alpha cuts is $\gamma(A) = 0.5904$ whereas for the FS \textit{B} $\gamma(B)$ is \(0.9578\). Note that as expected, for the FS \textit{A} the agreement is considerably smaller than for FS \textit{B} since its shape is \textit{sharper} and \textit{shorter} due to the differences in the weighted comparisons of alpha-cut lengths.

\subsection{Synthetic Example 2 using Gaussian FSs}
\begin{figure}[h]
\includegraphics[scale=0.4]{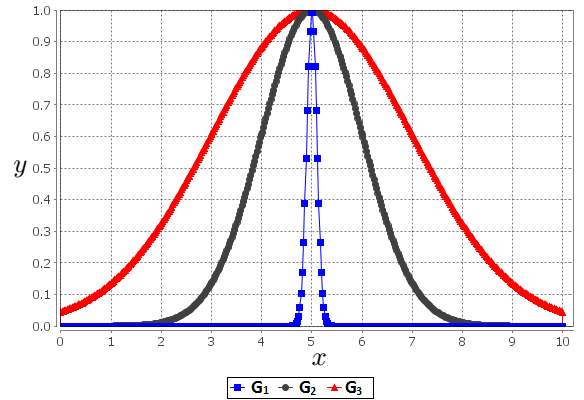}
\centering
	\caption{{\fontsize{8pt}{1em} \selectfont Three Fuzzy sets with Gaussian membership functions.}}
	\label{fig:normal}
\end{figure}

Consider the FSs $G_{1}$,$G_{2}$ and $G_{3}$ with Gaussian membership functions depicted in Fig. \ref{fig:normal}. These FSs are both normal and convex which, from the perspective of the assumptions made in order to develop the measure, indicates that all of the intervals/participants have agreed in a region. They have been arbitrarily chosen to have the same mean ($m=5$) but different standard deviations ($\sigma_{1}=0.1$, $\sigma_{2}=1.0$, $\sigma_{3}=2.0$). Again, no assumptions about the method employed to generate them have been made.
By using 10 alpha cuts, the calculated agreement ratio for the three FSs is $\gamma(G_1)\approx \gamma(G_2)\approx \gamma(G_3)\approx0.6518$ and similar results can be found by using different numbers of alpha-cuts. Further discussion of these results are found on Section \ref{sec:disc}.

\subsection{Synthetic Example 1 using IAA generated FS}
\begin{figure}[h]
	\includegraphics[scale=0.35]{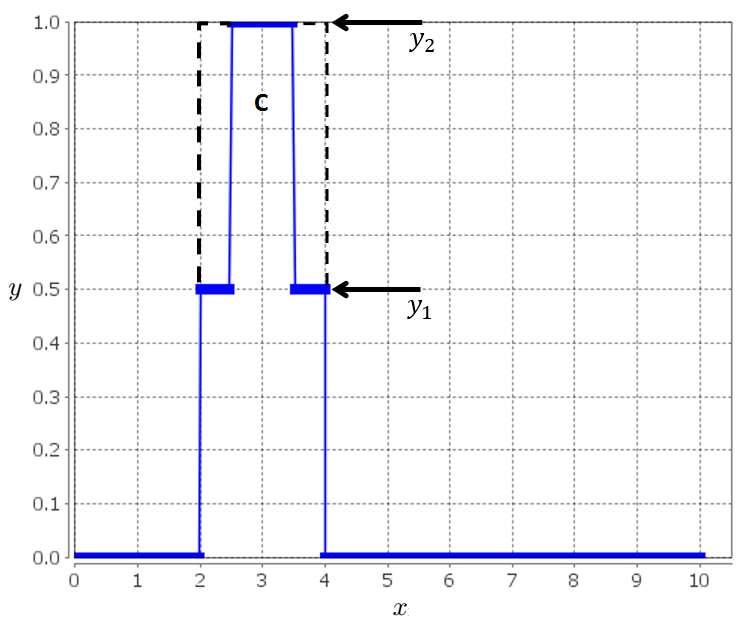}
	\centering
	\caption{{\fontsize{8pt}{1em} \selectfont Fuzzy set generated from 2 intervals}}
	\label{fig:fsA}
\end{figure}

Consider the FS $C$ depicted in Fig. \ref{fig:fsA} generated by two intervals \({ \bar { C }  }_{ 1 }=\left[ 2,4 \right] \) and \({ \bar { C }  }_{ 2 }=\left[ 2.5,3.5 \right] \). The agreement ratio $ \gamma (C)$ is calculated by dividing the total length of the union of all combinations of intervals where there is at least an intersection of 2 intervals ($y_2$) by the union ($y_1$) of all intervals.
\[\gamma (C)=1\left( \frac { 1 }{ 2 }  \right) /1=0.5\]
Note that in this example with 2 intervals, it can be simply considered as the length of the intersection of both intervals divided by the length of the union.

\subsection{Synthetic Example 2 using IAA (non-convex)}
\begin{figure}[h]
	\includegraphics[scale=0.35]{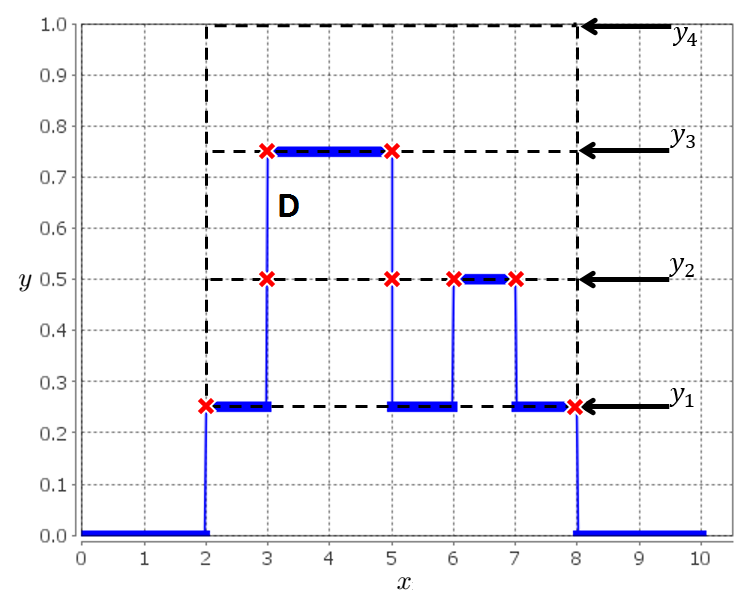}
	\centering
	\caption{{\fontsize{8pt}{1em} \selectfont Fuzzy set generated from 4 intervals}}
	\label{fig:fsB}
\end{figure}

Consider the FS $A$ depicted in Fig. \ref{fig:fsB} generated by the intervals \({ \bar { D }  }_{ 1 }=\left[ 2,5 \right] \) and \({ \bar { D }  }_{ 2 }=\left[ 3,5 \right] \), \({ \bar { D }  }_{ 3 }=\left[ 6,8 \right] \) and \({ \bar { D }  }_{ 4 }=\left[ 3,7 \right] \). The agreement ratio $ \gamma (D)$ is obtained by adding the weighted (${y}_{4}$, ${y}_{3}$ and ${y}_{2}$)  similarities between the lengths of the union of combinations of intervals where there is at least 4 and 3, 3 and 2, and 2 and 1 intervals respectively:
\[\gamma (D)=\frac { 1\left( \frac { 0 }{ 2 }  \right) +\cfrac { 3 }{ 4 } \left( \frac { 2 }{ 3 }  \right) +\cfrac { 2 }{ 4 } \left( \frac { 3 }{ 6 }  \right)  }{ 1+\cfrac { 3 }{ 4 } +\cfrac { 2 }{ 4 }  } =0.333\]

As can be seen in the above example (marked with red crosses), the length at the $y_4$ level is $0$, at the $y_3$ level is $2$, at the $y_2$ level is $3$ and at the $y_1$ level is $6$. Note that at the $y_2$ level there are 2 intervals which have to be added.
\begin{figure}[h]
	\includegraphics[scale=0.4]{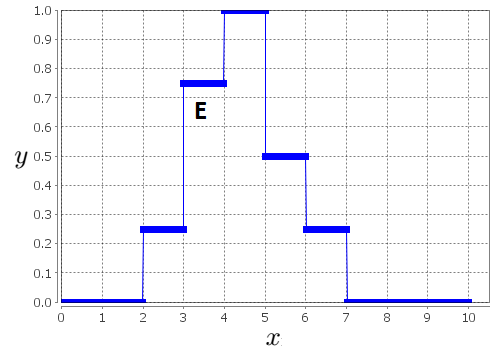}
	\centering
	\caption{{\fontsize{8pt}{1em} \selectfont Fuzzy set $E$ generated from 4 intervals}}
	\label{fig:fsE}
\end{figure}
Now lets consider the FS $E$ depicted in Fig. \ref{fig:fsE} created using the next intervals: \({ \bar { E }  }_{ 1 }=\left[ 2,5 \right] \) and \({ \bar { E }  }_{ 2 }=\left[ 3,5 \right] \), \({ \bar { E }  }_{ 3 }=\left[ 4,6 \right] \) and \({ \bar { E }  }_{ 4 }=\left[ 3,7 \right] \). Note that they are almost the same intervals than for FS $D$ but except one and such difference allows the generated FS to have a region with total agreement, i.e., the interval $ [4,5] $ . As such, we can expect a higher agreement ratio when comparing $\gamma(E)$ to $\gamma(D)$. Calculations using (\ref{eq:agrRat}) and (\ref{eq:length}) are shown below:

$\gamma (E)=\frac { 1\left( \frac { 1 }{ 2 }  \right) +\cfrac { 3 }{ 4 } \left( \frac { 2 }{ 3 }  \right) +\cfrac { 2 }{ 4 } \left( \frac { 3 }{ 5 }  \right)  }{ 1+\cfrac { 3 }{ 4 } +\cfrac { 2 }{ 4 }  } =\frac { 4 }{ 7 } \approx 0.5778$

\subsection{Application to TESS Data using the IAA} \label{sec:ex1}
In our previous work \cite{Navarro2016a}, we described a process of interval-valued data collection from different groups of people involved in assessment of function following sarcoma surgery, namely: Patients, Physiotherapists, Surgeons and a fourth one created from the combined responses from both Physiotherapists and Surgeons (PS) which together represent the body of ``medical professionals''.

We surveyed thirty-six participants (12 sarcoma surgeons, 13 physiotherapists and 12 patients undergoing lower limb salvage surgery) on 5 linguistic terms used to describe the extent of difficulty to perform daily activities: ``\textit{impossible to do}'',``\textit{extremely difficult}'', ``\textit{moderately difficult}'', ``\textit{a little bit difficult}'', and ``\textit{not at all difficult}''. Subsequently, we used the gathered intervals in order to generate T1 FSs by using the IAA and calculated basic FS attributes and their respective agreement ratio using (\ref{eq:agrRat}) and Algorithm \ref{alg:alpha}  (see Table \ref{tab:tessAgrRatio}).

\begin{figure}[!ht]
\centering
	\subfloat[\label{subfig:patient}]{
		\includegraphics[width=0.35\textwidth]{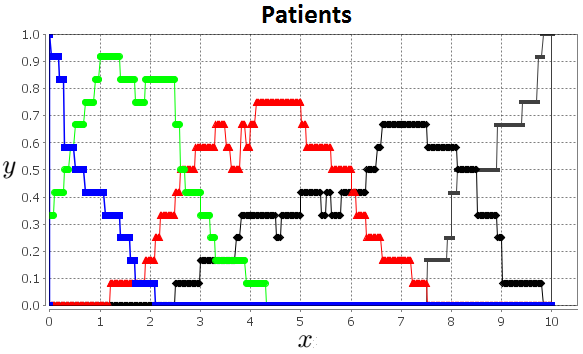}
	}\hfill
	\subfloat[\label{subfig:physio}]{
		\includegraphics[width=0.35\textwidth]{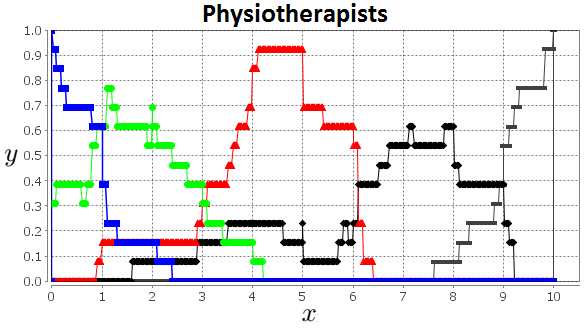}
	}\hfill
    \subfloat[\label{subfig:surgeon}]{
		\includegraphics[width=0.35\textwidth]{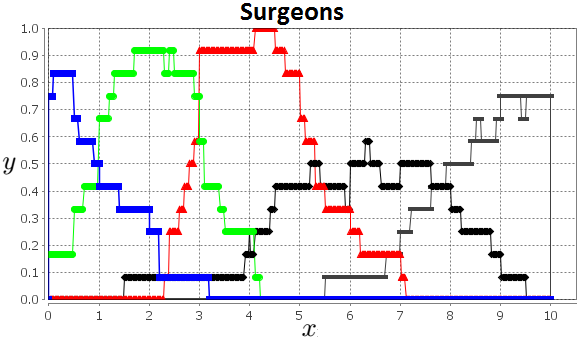}
	}\hfill\subfloat[\label{subfig:all}]{
		\includegraphics[width=0.35\textwidth]{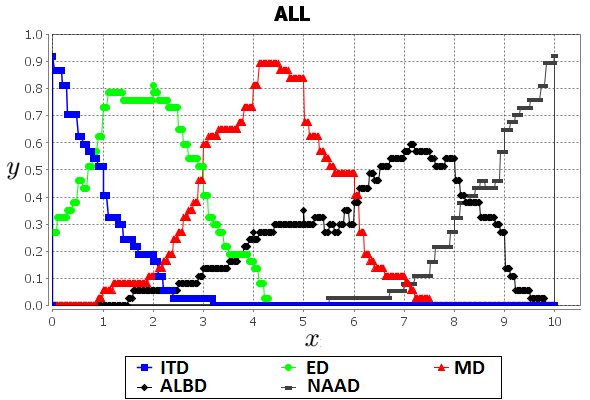}
	}\hfill
	\caption{FSs modelling the word concepts from left to right: \textit{ITD}, \textit{ED}, \textit{MD}, \textit{ALBD}, and \textit{NAAD}, generated from different sources: (a) is for patients. (b) is for Physiotherapists. (c) is for Surgeons. (d) is for the combined responses from all groups.}
\label{fig:wordsConceptsFS}
\end{figure}
Figure \ref{fig:wordsConceptsFS} depicts the FSs for the inter-patient agreement for the 5 linguistic descriptions (words) and different groups. Note that at first sight, by considering the width and height of the FSs, the term \textit{A little bit difficult} is the most subjective and less accepted among the different groups of stakeholders. Moreover, the agreement ratio ($\gamma$) substantially facilitates performing judgements about the acceptance of the different linguistic terms in a given scenario.


\begin{table}[htbp]
  \centering
  \caption{Agreement ratios in the context of other statistical measures}
    \begin{tabular}{ccccc}
    \toprule
    \textbf{Group} & \multicolumn{1}{c}{\textbf{Ling. Term}} & \textbf{Height} & \textbf{Centroid} & \textbf{Agr. Rat.} \\
    \midrule
    \multirow{5}[0]{*}{Patient} & ITD   & \textbf{1.000} & 0.686 & 0.669 \\
          & ED    & 0.917 & 1.711 & 0.652 \\
          & MD    & 0.750 & 4.356 & 0.433 \\
          & ALBD  & 0.667 & 6.433 & 0.323 \\
          & NAAD  & \textbf{1.000} & 9.051 & 0.775 \\
          \midrule
    \multirow{5}[0]{*}{Physiotherapist} & ITD   & \textbf{1.000} & 0.727 & 0.629 \\
          & ED    & 0.769 & 1.767 & 0.368 \\
          & MD    & 0.923 & 4.312 & 0.733 \\
          & ALBD  & 0.615 & 6.462 & 0.242 \\
          & NAAD  & \textbf{1.000} & 9.279 & 0.701 \\
          \midrule
    \multirow{5}[0]{*}{Surgeon} & ITD   & 0.917 & 0.988 & 0.554 \\
          & ED    & 0.917 & 2.085 & 0.691 \\
          & MD    & \textbf{1.000} & 4.289 & \textbf{0.803} \\
          & ALBD  & 0.583 & 6.126 & 0.185 \\
          & NAAD  & 0.833 & 8.555 & 0.461 \\
          \midrule
    \multirow{5}[0]{*}{ALL} & ITD   & 0.919 & 0.817 & 0.687 \\
          & ED    & 0.811 & 1.862 & 0.559 \\
          & MD    & 0.892 & 4.319 & 0.717 \\
          & ALBD  & 0.595 & 6.346 & 0.280 \\
          & NAAD  & 0.919 & 8.901 & 0.719 \\
    \bottomrule
    \end{tabular}%
  \label{tab:tessAgrRatio}%
\end{table}%

From Table \ref{tab:tessAgrRatio}, it is worthwhile to note that the linguistic terms \textit{A little bit difficult} and \textit{Moderately difficult} from \textit{Surgeons} are contrastingly the less and most agreed, respectively. Such information is not inferred from other FS measures (e.g., height, centroid, support size) and can help to analyse through a numerical approach the aptness of different linguistic terms in specific contexts.

\section{Discussion}\label{sec:disc}
The proposed Agreement Ratio was developed as a means of having a tool of analysis for deciding which linguistic descriptors for patients' conditions can be more apt for a given scenario. As previously mentioned, it is of a key importance to know how similar the perception of the meaning of a given word is across different stakeholders while taking into account participants' uncertainty represented through intervals.
We have shown results considering synthetic examples with FSs using some of the most used types of membership function and a data-driven approach (IAA) in which it can be highlighted that:
\begin{enumerate}
\item The application of the proposed measure produces meaningful results when using FSs which express both, inter-source uncertainty in the domain and agreement in the membership.
\item The measure does not consider the whole scale being surveyed, but only the function domain and the proportional changes in ratio from the membership function support to the top (highest agreement). Therefore, the application of the proposed measure to Gaussian FSs produces similar values due to the Gaussian shape "smoothness" scaled through different function supports.
\end{enumerate}
Regarding the application of the measure to the data obtained from the survey on TESS linguistic descriptors, we acknowledge that this is a limited sample size but if low agreement values still being obtained for \textit{ED} and \textit{ALBD} from a larger sample, then this may suggest that there is a potential risk of miscommunication in this scenario. Consequently, a set of different (and more unanimously understood) linguistic expressions could be seek to replace the current ones.

\section{Conclusions and Future Work} \label{sec:conc}
In this study, we proposed a simple method to obtain an agreement ratio focused on inter-participant agreement through FSs generated from a data-driven approach, namely the IAA. We provided synthetic examples to show the calculations and also the results of the measure on a real world dataset obtained from different groups of people involved in a medical assessment scenario in which perceptions are key.
The results show that the proposed measure can provide directly a means of evaluating the aptness of a Fuzzy Set representing a word in a given group over others. This measure has an important potential in several medical-patient intercommunication scenarios in which differences in background and context may produce misleading /assessments interpretations among different groups.\\
 We foresee the proposed measure usefulness in practical scenarios in which decision based on linguistic assessments are needed. For example, it can be useful to analyse a codebook with potential linguistic terms as candidates in which it is needed to avoid ambiguity as much as possible, e.g., by grouping/ranking similar terms using a defined criterion (centroid, etc.) and selecting those with the highest agreement ratio $\gamma$. Another application can be to use the agreement ratio to measure the level of consensus and allow discussion of the results among the stakeholders and repeat the survey process until more considerable agreement ratios are obtained.
Although the measure proposed in this paper has only been designed for T1 FSs, we have already explored the extension of the measure to T2 FSs which will be presented in a future publication. The extension  is focused on enabling the application of the measure to other common FS generation techniques which generate T2 FSs (e.g., the Enhanced Interval Approach). We also plan to develop a more detailed methodology for the selection of words for CWW engines based on the proposed agreement ratio and explore its results in comparison to other approaches.
\section*{Acknowledgements}
We would like to acknowledge the support from Josie McCulloch for all her beneficial writing assistance and proof reading on the paper.
\bibliography{references}

\begin{thebibliography}{10}
\providecommand{\url}[1]{#1}
\csname url@samestyle\endcsname
\providecommand{\newblock}{\relax}
\providecommand{\bibinfo}[2]{#2}
\providecommand{\BIBentrySTDinterwordspacing}{\spaceskip=0pt\relax}
\providecommand{\BIBentryALTinterwordstretchfactor}{4}
\providecommand{\BIBentryALTinterwordspacing}{\spaceskip=\fontdimen2\font plus
\BIBentryALTinterwordstretchfactor\fontdimen3\font minus
  \fontdimen4\font\relax}
\providecommand{\BIBforeignlanguage}[2]{{%
\expandafter\ifx\csname l@#1\endcsname\relax
\typeout{** WARNING: IEEEtran.bst: No hyphenation pattern has been}%
\typeout{** loaded for the language `#1'. Using the pattern for}%
\typeout{** the default language instead.}%
\else
\language=\csname l@#1\endcsname
\fi
#2}}
\providecommand{\BIBdecl}{\relax}
\BIBdecl

\bibitem{Davis1996}
A.~M. Davis, J.~G. Wright, J.~I. Williams, C.~Bombardier, A.~Griffin, and R.~S.
  Bell, ``{Development of a measure of physical function for patients with bone
  and soft tissue sarcoma},'' \emph{Quality of Life Research}, vol.~5, no.~5,
  pp. 508--516, 1996.

\bibitem{Zadeh1996}
L.~A. Zadeh, ``{Fuzzy logic equals Computing with words},'' \emph{IEEE
  Transactions on Fuzzy Systems}, vol.~4, no.~2, pp. 103--111, 1996.

\bibitem{Herrera2009}
F.~Herrera, S.~Alonso, F.~Chiclana, and E.~Herrera-Viedma, ``{Computing with
  words in decision making: Foundations, trends and prospects},'' \emph{Fuzzy
  Optimization and Decision Making}, vol.~8, pp. 337--364, 2009.

\bibitem{Wagner2014}
C.~Wagner, S.~Miller, J.~Garibaldi, D.~Anderson, and T.~Havens, ``{From
  Interval-Valued Data to General Type-2 Fuzzy Sets},'' \emph{IEEE Transactions
  on Fuzzy Systems}, vol.~23, pp. 248--269, 2014.

\bibitem{Coupland2010}
S.~Coupland, J.~M. Mendel, and D.~Wu, ``{Enhanced Interval Approach for
  encoding words into interval type-2 fuzzy sets and convergence of the word
  FOUs},'' in \emph{International Conference on Fuzzy Systems}.\hskip 1em plus
  0.5em minus 0.4em\relax IEEE, 2010, pp. 1--8.

\bibitem{Feilong2008a}
L.~Feilong and J.~M. Mendel, ``{Encoding Words Into Interval Type-2 Fuzzy Sets
  Using an Interval Approach},'' \emph{Fuzzy Systems, IEEE Transactions on},
  vol.~16, no.~6, pp. 1503--1521, 2008.

\bibitem{Setnes1998}
M.~Setnes, R.~Babuska, U.~Kaymak, and H.~R. van Nauta~Lemke, ``{Similarity
  measures in fuzzy rule base simplification.}'' \emph{IEEE transactions on
  systems, man, and cybernetics. Part B, Cybernetics : a publication of the
  IEEE Systems, Man, and Cybernetics Society}, vol.~28, no.~3, pp. 376--86,
  1998.

\bibitem{Jaccard1908}
P.~Jaccard, ``{Nouvelles recherches sur la distribution florale},''
  \emph{Bulletin de la Soci{\'{e}}t{\'{e}} vaudoise des sciences naturelles},
  vol. Volume 5, no. 163, 1908.

\bibitem{Wagner2013b}
\BIBentryALTinterwordspacing
C.~Wagner, S.~Miller, and J.~M. Garibaldi, ``{Similarity based applications for
  data-driven concept and word models based on type-1 and type-2 fuzzy sets},''
  \emph{2013 IEEE International Conference on Fuzzy Systems (FUZZ-IEEE)}, pp.
  1--9, 7 2013. [Online]. Available:
  \url{http://ieeexplore.ieee.org/lpdocs/epic03/wrapper.htm?arnumber=6622466}
\BIBentrySTDinterwordspacing

\bibitem{Navarro2016a}
J.~Navarro, C.~Wagner, U.~Aickelin, L.~Green, and R.~Ashford, ``{Exploring
  Differences in Interpretation of Words Essential in Medical Treatment by
  Patients and Medical Professionals},'' in \emph{IEEE World Congress On
  Computational Intelligence}, Vancouver, Canada, 2016.

\bibitem{Miller2014}
S.~Miller, C.~Wagner, and J.~M. Garibaldi, ``{Exploring Statistical Attributes
  Obtained from Fuzzy Agreement Models},'' in \emph{IEEE International
  Conference on Fuzzy Systems}, 2014.

\bibitem{zadeh1975concept}
L.~A. Zadeh, ``{The concept of a linguistic variable and its application to
  approximate reasoning-I},'' \emph{Information sciences}, vol.~8, no.~3, pp.
  199--249, 1975.

\bibitem{Miller2012}
S.~Miller, C.~Wagner, J.~M. Garibaldi, and S.~Appleby, ``{Constructing general
  type-2 fuzzy sets from interval-valued data},'' \emph{IEEE International
  Conference on Fuzzy Systems}, pp. 10--15, 2012.

\bibitem{Wagner2011}
C.~Wagner and H.~Hagras, ``{Employing zSlices Based General Type-2 Fuzzy Sets
  to Model Multi Level Agreement},'' \emph{IEEE SSCI 2011: Symposium Series on
  Computational Intelligence - T2FUZZ 2011: 2011 IEEE Symposium on Advances in
  Type-2 Fuzzy Logic Systems}, pp. 50--57, 2011.

\bibitem{Wagner2010a}
------, ``{Toward General Type-2 Fuzzy Logic Systems Based on zSlices},''
  \emph{Fuzzy Systems, IEEE Transactions on}, vol.~18, no.~4, pp. 637--660,
  2010.

\bibitem{Berry2003}
D.~C. Berry, D.~K. Raynor, P.~Knapp, and E.~Bersellini, ``{Patients'
  understanding of risk associated with medication use: impact of European
  Commission guidelines and other risk scales.}'' \emph{Drug safety : an
  international journal of medical toxicology and drug experience}, vol.~26,
  no.~1, pp. 1--11, 2003.

\end{thebibliography}
\end{document}